\documentclass[10pt,twocolumn,letterpaper]{article}

\usepackage[usenames,dvipsnames]{color}
\usepackage{caption}
\usepackage[labelformat=simple,labelfont={}]{subcaption}

\usepackage{cvpr}
\usepackage{times}
\usepackage{epsfig}
\usepackage{graphicx}
\usepackage{amsmath}
\usepackage{amssymb}
\usepackage{xr} 

\newcommand{\Ls}{\mathcal{L}}

\newcommand{\W}{\mathbf{W}}

\usepackage{url}
\usepackage{dcolumn}
\usepackage[applemac]{inputenc}
\usepackage{verbatim}
\usepackage{rotating}
\usepackage{url}
\usepackage{amsfonts}
\usepackage[nice]{nicefrac}
\usepackage{float}
\usepackage{mathrsfs}
\usepackage{pgf}
\usepackage{slashbox}
\usepackage{setspace}

\usepackage[breaklinks=true,bookmarks=false]{hyperref}

\cvprfinalcopy 

\def\cvprPaperID{1927} 

\newif\ifcomments

\commentsfalse

\ifcomments
\newcommand{\comments}[1]{#1}
\else
\newcommand{\comments}[1]{}
\fi

\newcommand{\todo}[1]{\comments{\textcolor{red}{[todo: #1]}}}
\newcommand{\maybe}[1]{\comments{\textcolor{Dandelion}{[maybe: #1]}}}

\definecolor{jbyclr}{rgb}{.9,0,.9}

\begin{document}

\title{Recombinator Networks: Learning Coarse-to-Fine Feature Aggregation}

\author{Sina Honari${^1}$, Jason Yosinski${^2}$, Pascal Vincent${^{1,4}}$, Christopher Pal${^3}$\\
 ${^1}$University of Montreal, ${^2}$Cornell University, ${^3}$Ecole Polytechnique of Montreal, ${^4}$CIFAR\\
{\tt\small ${^1}$\{honaris, vincentp\}@iro.umontreal.ca, ${^2}$yosinski@cs.cornell.edu, ${^3}$christopher.pal@polymtl.ca}
}

\maketitle

\begin{abstract}

Deep neural networks with alternating convolutional, max-pooling and decimation layers are widely used in state of the art architectures for computer vision.
Max-pooling purposefully discards precise spatial information in order to create features that are more robust, and typically organized as lower resolution spatial feature maps. On some tasks, such as whole-image classification, max-pooling derived features are well suited; however, for tasks requiring precise localization, such as pixel level prediction and segmentation, max-pooling destroys exactly the information required to perform well.
Precise localization may be preserved by shallow convnets without pooling but at the expense of robustness. 
Can we have our max-pooled multi-layered cake and eat it too?
Several papers have proposed summation and concatenation based methods for combining upsampled coarse, abstract features with finer features to produce robust pixel level predictions. Here we introduce another model --- dubbed Recombinator Networks --- where coarse features inform finer features early in their formation such that finer features can make use of several layers of computation in deciding how to use coarse features.
The model is trained once, end-to-end  and performs better than summation-based architectures, reducing the error from the previous state of the art on two facial keypoint datasets, AFW and AFLW, by 30\% and beating the current state-of-the-art on 300W without using extra data. We improve performance even further by adding a denoising prediction model based on a novel convnet formulation.
\end{abstract}
\begin{figure*}[!htb]
\centering
\includegraphics[width=.47\textwidth]{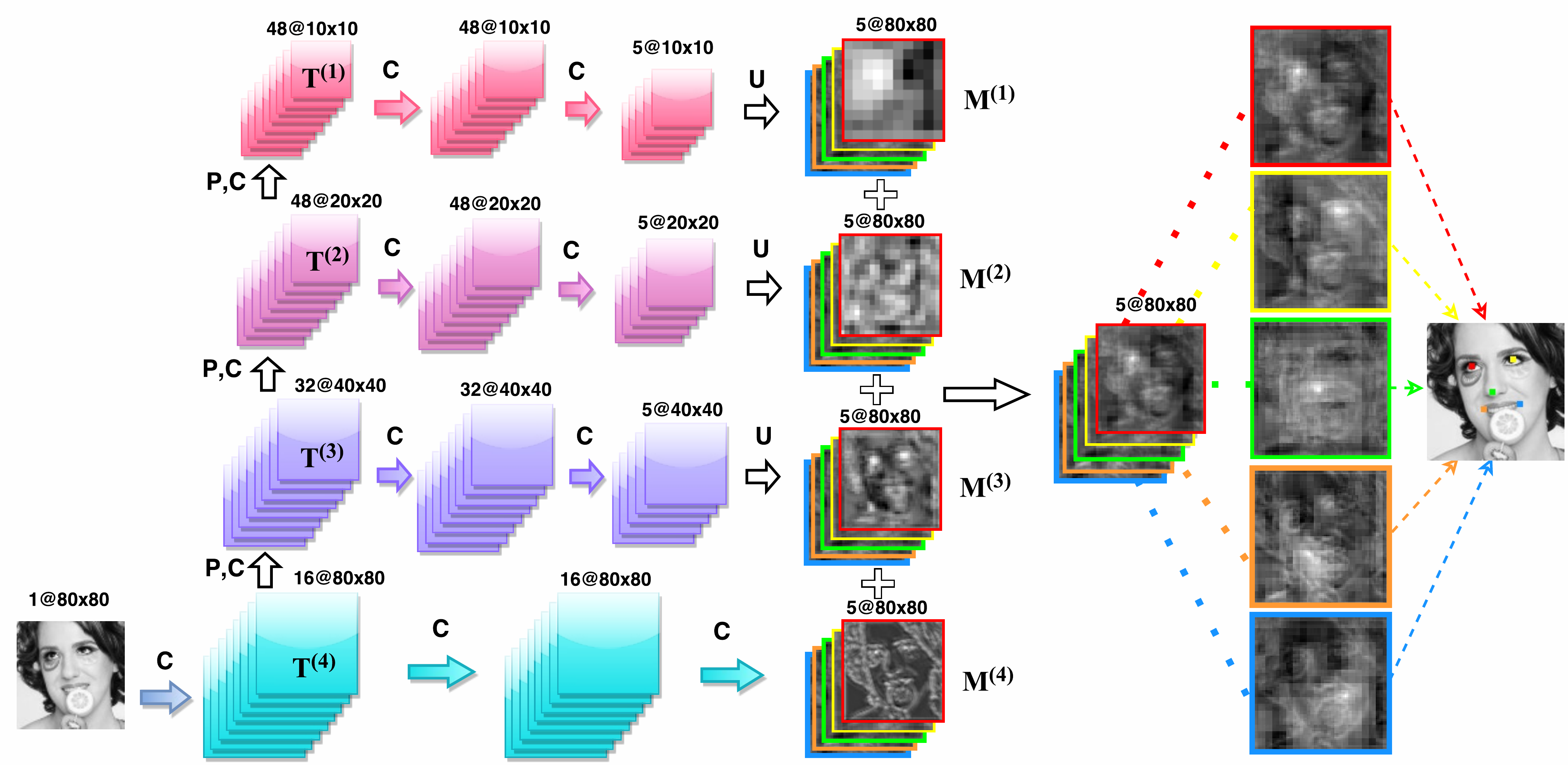}
\includegraphics[width=.52\textwidth]{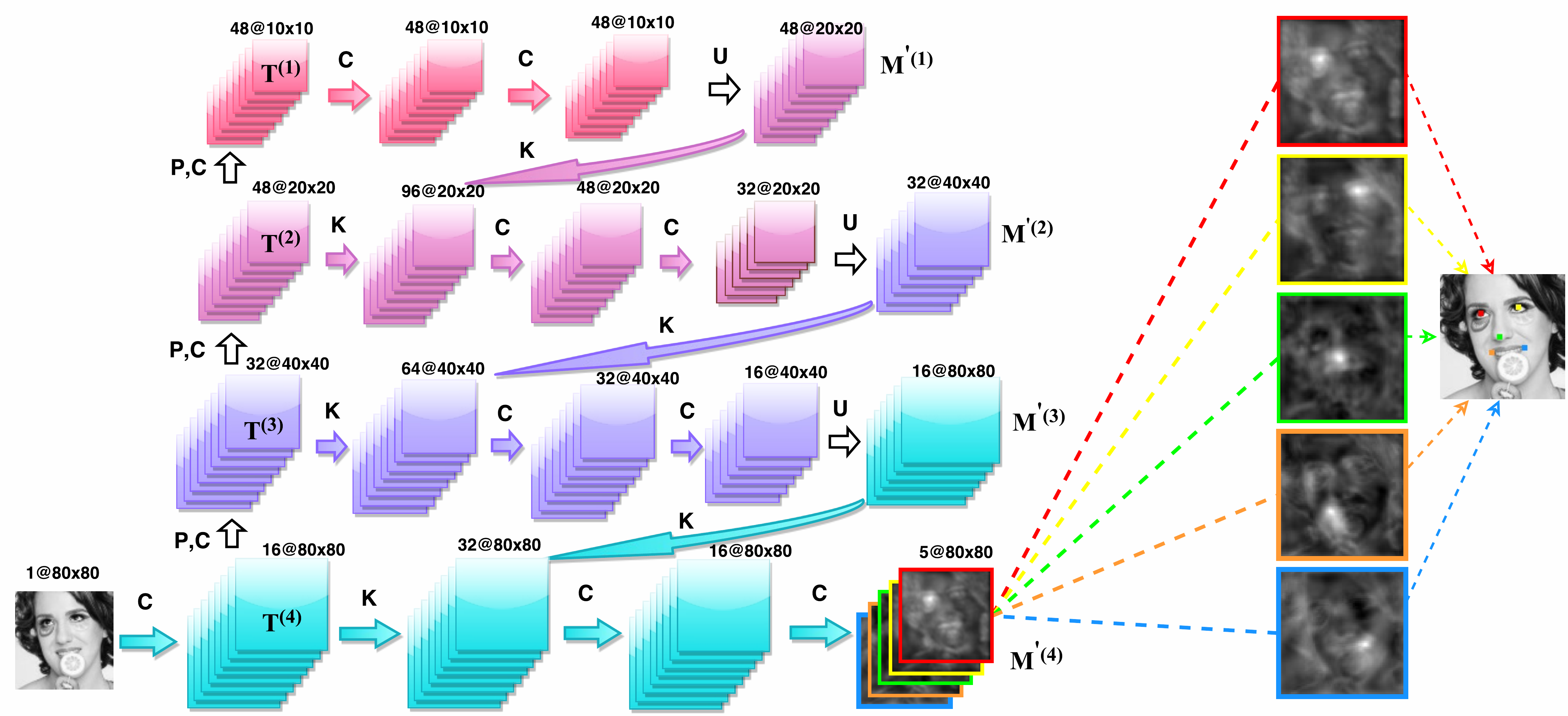}
\caption{(\emph{Left}) Architecture of summation based coarse-fine network (SumNet). C is a convolutional layer. P,C represents a pooling layer followed by a convolutional layer.
All convolutions are $3 \times 3$ and all poolings are $2 \times 2$.
All convolutional layers are followed by ReLU non-linearity except the last convolutional layer in each branch.
U represents an upsampling layer. Each branch's output is 5 feature maps of size $80\times80$.
FCN/Hypercolumn models use this architecture.
(\emph{Right}) Architecture of the Recombinator Networks (RCN). All convolutions are $3 \times 3$ and all poolings are $2 \times 2$.
All upsamplings are by a factor of 2. $K$ represents concatenation of two sets of feature maps along the feature map dimension.
All convolutional layers are followed by ReLU non-linearity except the one right before the softmax.
In the Recombinator Networks model with skip connections (not shown), each branch takes upsampled features not only from one coarser branch, but from all coarser branches.
}
\vspace{-2ex}
\label{fig:mscale}
\end{figure*}
\vspace{-2ex}
\section{Introduction}

Recent progress in computer vision has been driven by the use of large
convolutional neural networks. Such networks benefit from alternating convolution and
pooling layers \cite{kriz2012, sermanet2011traffic, sermanet2013pedestrian, szegedy2014going, simonyan2014very, sun2014deep, zhang2015learning}
where the pooling layers serve to summarize small regions
of the layer below.
The operations of convolution, followed by max-pooling, then decimation cause features in subsequent layers of the network
to be increasingly 
translation invariant, more robust, and to more coarsely summarize progressively larger regions
of the input image.
As a result, features in the fourth or fifth convolutional layer serve as more robust detectors of more global, but spatially imprecise high level patterns like text or human faces \cite{yosinski-2015-ICML-DL-understanding-neural-networks}. In practice these properties are
critical for many visual tasks, and they have been particularly
successful at enabling whole image classification \cite{kriz2012, szegedy2014going, simonyan2014very}.
However, for other types of vision tasks these architectural elements are not as well suited. For example on tasks requiring pixel-precise localization or labeling,
features arising from max-pooling and decimation operations can only provide approximate localization, as in the process of
creating them, the network has already thrown out precise spatial information by design.
%
%
If we wish to generate features that preserve accurate localization, we may do so using shallow networks without max-pooling,
but shallow networks without pooling
cannot learn robust, invariant features.
What we would like is to have our cake and eat it too: to combine the best of both worlds, merging finely-localized information from
shallow, non-pooled networks with robust, coarsely-localized features computed by deep, pooled networks.

Several recently proposed approaches \cite{long2014fully,hariharan2014hypercolumns,tompson2014efficient} address this by adding or concatenating the features obtained across multiple levels. We use this approach in our baseline model termed \emph{SumNet} for our task of interest: facial keypoint localization. 
To the best of our knowledge this is the first time this general approach has been applied to the problem of facial keypoint localization and even our baseline is capable of yielding state of the art results.
A possible weakness of these approaches however is that all detection paths, from coarsely to finely localized features, only become aggregated at the very end of the feature processing pipeline.
As a thought experiment to illustrate this approach's weakness, imagine that we 
have a photo of a boat floating in the ocean and would like to train a convnet to predict with single pixel accuracy a keypoint corresponding to the tip of the boat's bow. Coarsely localized features\footnote{From now on we use the shorthand fine/coarse features to mean finely/coarsely localized features.} could highlight the rough region of the bow of the boat, and finely localized features could be tuned to find generic boat edges, but the fine features must remain generic, being forced to learn boat edge detectors for all possible ocean and boat color combinations. This would be difficult, because boat and ocean pixels could take similar colors and textures. Instead, we would like a way for the coarse features which contain information about the global scene structure (perhaps that the water is dark blue and the boat is bright blue) to provide information to the fine feature detectors earlier in their processing pipeline.
Without such information, the fine feature detectors would be unable to tell which half of a light blue/dark blue edge was ocean and which was boat. In the \emph{Recombinator Networks} proposed in this paper, the finely localized features are conditioned on higher level more coarsely localized information. It results in a model which is deeper but -- interestingly -- trains faster than the summation baseline and yields more precise localization predictions.
In summary, this work makes the following contributions:
\begin{enumerate}
\item \vspace{-1.7ex} We propose a novel architecture --- the Recombinator Networks --- for combining information over different spatial localization resolutions (Section~\ref{cf-models}).
\item \vspace{-1.7ex} We show how a simple denoising model may be used to enhance model predictions (Section~\ref{sec:denoising_model}).
\item \vspace{-1.7ex} We provide an in-depth empirical evaluation of a wide variety of relevant architectural variants (Section~\ref{sec:evaluation}).
\item \vspace{-1.7ex} We show state of the art performance on two widely used and competitive evaluations for facial keypoint localization 
(Section \ref{ex:compare}).
\end{enumerate}

\section{Related work}
\label{related-work}

\textbf{Keypoint localization methods:} 
Our task of interest is the well studied problem of facial keypoint localization 
\cite{
zhu2015face, zhang2015learning, yu2013, asthana2013robust, tzimiropoulos2014gauss, zhang2014coarse, cao2014, xiong2013, cao2014face, ren2014face, zhu2012} illustrated in Figure~\ref{fig:mscale}.
Precise facial keypoint localization is often an essential preprocessing step
for face recognition \cite{asthana2011fully}
and detection \cite{zhu2012}.
Recent face verification models like
DeepFace \cite{taigman2014deepface} and DeepID2 \cite{sun2014deep} also include keypoint
localization as the first step.
There have been many other approaches to general keypoint localization, including active appearance models
\cite{cootes2001active, zhao2013locality}, constrained local models \cite{cristinacce2006feature, saragih2009face, cristinacce2008automatic, asthana2013robust},
active shape models \cite{cristinacce2007boosted}, 
point distribution models \cite{cootes1995active}, structured model prediction \cite{baltruvsaitis2014continuous, tompson2014efficient},
tree structured face models \cite{zhu2012}, group sparse learning based methods \cite{yu2013pose},
shape regularization models that combines multiple datasets \cite{smith2014collaborative},
feature voting based landmark localization \cite{smith2014nonparametric, yang2013sieving} and convolutional neural networks
based models \cite{zhang2014facial, sun2013deep, zhang2015learning}.
Two other related models are \cite{tompson2014efficient}, where a multi-resolution model is proposed with dual coarse/fine paths and tied filters, and \cite{sun2013deep}, which uses a cascaded architecture to refine predictions over several stages.
Both of these latter models make hard decisions using coarse information halfway through the model.

\textbf{Approaches that combine features across multiple levels:} 
Several recent models --- including
the fully convolutional networks (FCNs) in \cite{long2014fully}, the Hypercolumn model \cite{hariharan2014hypercolumns}, and the localization model of Tompson et al. \cite{tompson2014efficient} --- generate features or predictions at multiple resolutions, upsample the coarse features to the fine resolution, and then
add or concatenate the features or predictions together.
This approach has generally worked well, improving on previous state of the art results in
detection, segmentation, and human-body pose estimation \cite{hariharan2014hypercolumns,long2014fully,tompson2014efficient}.
In this paper we create a baseline model similar to these approaches that we refer to as \emph{SumNet} in which we use a network that aggregates information from features across different levels in the hierarchy of a conv-pool-decimate network using concatenation followed by a weighted sum over feature maps prior to final layer softmax predictions. Our goal in this paper is to improve upon this architecture.
Differences between the Recombinator Networks and related architectures are summarized in Table~\ref{tab:models}.
U-Net \cite{ronneberger2015u} is another model that merges features across multiple levels and has a very similar architecture to Recombinator Networks. The two models have been developed independently and were designed for different problems\footnote{For keypoint localization, we apply the softmax \emph{spatially} i.e. across possible \emph{spatial locations}, whereas for segmentation \cite{hariharan2014hypercolumns, long2014fully, ronneberger2015u} it is applied across all possible \emph{classes} for each pixel.}.
Note that none of these models use a learned denoising post-processing as we do (see section \ref{sec:denoising_model}).

\section{Summation versus Recombinator Networks}
\label{cf-models}
In this section we describe our baseline SumNet model based on a common architectural design where information from different levels of granularity are merged just prior to predictions being made. We contrast this with the Recombinator Networks architecture.

\subsection{Summation based Networks}
The SumNet architecture, shown in Figure \ref{fig:mscale}(left), adds to the usual bottom to top convolution and spatial pooling, or ``trunk'',
a horizontal left-to-right ``branch'' at each resolution level. While spatial pooling progressively
reduces the resolution as we move ``up'' the network along the trunk, the horizontal branches only contains full
convolutions and element-wise non-linearities, with no spatial pooling, so that they can preserve
the spatial resolution at that level while doing further processing. 
The output of the finest resolution branch only goes through convolutional layers. The finest resolution layers keep positional information and use it 
to guide the coarser layers within the patch that they cannot have any preference, while the coarser resolution layers help finer layers to 
get rid of false positives.
The architecture then combines the rightmost low resolution output of all horizontal branches, into a single high resolution prediction, by first up-sampling\footnote{Upsampling can be performed either by tiling values or by using bilinear interpolation.
We found bilinear interpolation degraded performance in some cases, so we instead used the simpler tiling approach.} them all to the model's input image resolution ($80\times 80$ for our experiments)
and then taking a weighted sum to yield the pre-softmax values.
%
%
Finally, a softmax function is applied to yield the final location probability map for each keypoint.
%
Formally, given an input image $x$, define the \emph{trunk} of the network as a sequence of blocks of traditional groups of convolution, pooling and decimation operations. Starting from the layer yielding the coarsest scale feature maps we call the outputs of $R$ such blocks $T^{(1)}, \ldots, T^{(R)}$. At each level $r$ of the trunk we have a horizontal \emph{branch} that takes $T^{(r)}$ as its input and consists of a sequence of convolutional layers with no subsampling. The output of such a branch is a stack of $K$ feature maps, one for each of the $K$ target keypoints, at the same resolution as its input $T^{(r)}$, and we denote this output as $\mathrm{branch}(T^{(r)})$. It is then upsampled $\textrm{up}_{[\times F]}$ by some factor $F$ which returns the feature map to the original resolution of the input image.
Let these upsampled maps be $M^{(1)}_{1}, \ldots, M^{(R)}_{K}$ where $M^{(r)}_k$ is the score map given by the $r^{th}$ branch to the $k^{th}$ keypoint (left eye, right eye, $\ldots$). Each such map $M^{(r)}_k$ is a matrix of the same resolution as the image fed as input (i.e. $80 \times 80$). The score ascribed by branch $r$ for keypoint $k$ being at coordinate $i,j$ is given by $M^{(r)}_{k,i,j}$. The final \emph{probability map} for the location $Y_k$ of keypoint $k$ is given by a softmax over all possible locations. We can therefore write the model as
{\small
\begin{eqnarray}
&& M^{(1)} = \mathrm{up}_{[\times2^{R-1}]}( \mathrm{branch}(T^{(1)})) \nonumber \\
&& M^{(2)} = \mathrm{up}_{[\times2^{R-2}]}(\mathrm{branch}(T^{(2)})) \nonumber \\
&& \ldots \nonumber \\
&& M^{(R)} = \mathrm{branch}(T^{(R)}) \nonumber \\ 
&& P(Y_k | X = x) =
 \mathrm{softmax}\Big( \sum_{r=1}^R  \alpha_{rk} M^{(r)}_k \Big),
 \label{eqn:probmap}
\end{eqnarray}
}
%
%
%
where $\alpha_{rk}$ is a 2D matrix that gives a weight to every pixel location $i,j$ of keypoint $k$ in branch $r$.
The weighted sum of features over all branches taken here is
equivalent to concatenating the features of all branches and multiplying them in a set of weights, which results in
one feature map per keypoint.
This architecture is trained globally using gradient backpropagation to minimize the sum of negated conditional log probabilities of all $N$ training (input-image, keypoint-locations) pairs, for all $K$ keypoints $(x^{(n)},y_k^{(n)})$, with an additional regularization term for the weights
; i.e. we search for network parameters $W$ that minimize \footnote{
We also tried L2 distance cost between true and estimated keypoints (as a regression problem) and got worse results.
This may be due to the fact that a softmax probability map can be multimodal
, while L2 distance implicitly corresponds to likelihood of a \emph{unimodal} isotropic Gaussian.}
%
%
\vspace*{-0.5em}
{\small
\begin{align}
\label{eqn:training}
\Ls(\W) =  \frac{1}{N} \sum_{n=1}^{N} \sum_{k=1}^K -\log P(Y_k = y_k^{(n)} | X = x^{(n)}) + \lambda \| \mathcal{\W} \|^{2}.
\end{align}
\vspace*{-1.8em}
}
\subsection{The Recombinator Networks}
\label{sec:RCN}
In the SumNet model, different branches can only communicate through the updates received from the output layer and the features are merged linearly through summation.
%
In the Recombinator Networks (RCN) architecture, as shown in Figure \ref{fig:mscale}(right), instead of
taking a weighted sum of the upsampled feature maps
in each branch and then passing them to a softmax, the output of each branch is upsampled, then concatenated 
with the next level branch with one degree of finer resolution. In contrast to the SumNet model, each branch does not end in $K$ feature maps. The information stays in the form of a keypoint independent feature map. It is only at the end of the $R^{th}$ branch that feature maps are converted into a per-keypoint scoring representation that has the same resolution as the input image, on which a softmax is then applied.
As a result of RCN's different architecture,
branches pass more information to each other during training, 
such that convolutional layers in the finer branches get inputs from both coarse and fine layers,
letting the network learn how to combine them \textit{non-linearly}
to maximize the log likelihood of the keypoints given the input images.
%
The whole network is trained end-to-end by backprop. Following the previous conventions and by defining the concatenation operator on feature maps $A$, $B$ as $\mathrm{concat}(A,B)$, we can write the model as
\begin{small}
\begin{eqnarray}
&& M'^{(1)} = \mathrm{up}_{[\times2]}( \mathrm{branch}(T^{(1)} )) \nonumber \\
&& M'^{(2)} = \mathrm{up}_{[\times2]}(\mathrm{branch}(\mathrm{concat}(T^{(2)},M'^{(1)}))) \nonumber\\
&& \ldots \nonumber\\
&& M'^{(R)} = \mathrm{branch}(\mathrm{concat}(T^{(R)},M'^{(R-1)})) \nonumber
\\ 
&& P(Y_k|X=x) = \mathrm{softmax}(M'^{(R)}_k).
\end{eqnarray}
\end{small}
We also explore RCN with skip connections, where the features of each branch are concatenated with upsampled features of not only one-level coarser branch,
but all previous coarser branches and, therefore, the last branch computes
$M'^{(R)} = \mathrm{branch}(\mathrm{concat}(T^{(R)},M'^{(R-1)},M'^{(R-2)},\dots,M'^{(1)})$.
%
%
%
%
%
%
In practice, the information flow between different branches makes RCN converge faster and also perform better compared to the SumNet model.

\section{Denoising keypoint model}
\label{sec:denoising_model}

\begin{figure*}[h]
\begin{center}
\begin{subfigure}{0.60\textwidth}
\centering
\includegraphics[width=1\textwidth]{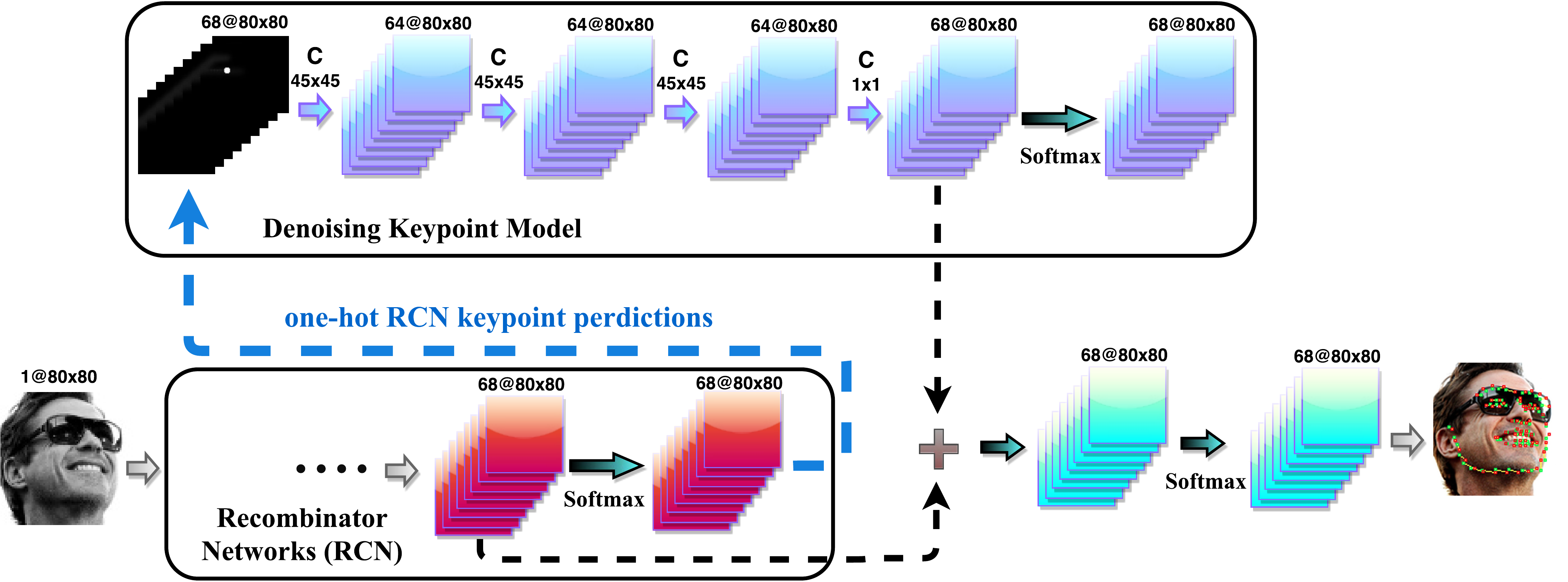}
\end{subfigure}
\end{center}
\caption{Denoising / joint keypoint model. The Recombinator Networks (RCN) and the keypoint location denoising models are trained
separately. At test time, the keypoint hard prediction of RCN is first injected into the denoising model
as one-hot maps. Then the pre-softmax values computed by the RCN and the denoising models are summed and pass through a final softmax
to predict keypoint locations.}
\label{fig:joint}
\end{figure*}

Convolutional networks are excellent edge detectors. If there are few samples with occlusion in the training sets,
 convnets have problem detecting occluded keypoints and instead select nearby edges (see some samples in Figures \ref{fig:cf_compare}, \ref{fig:occlusion}).
Moreover, the convnet predictions, especially on datasets with many keypoints,
do not always correspond to a plausible keypoint distribution 
and some keypoints jump off the curve (e.g. on the face contour or eye-brows)
irrespective of other keypoints' position (see some samples in Figure \ref{fig:compare_300W}).
This type of error can be addressed by using a 
structured output predictor on top of the convnet, that takes into account how likely the location of a keypoint is relative to other keypoints.
%
Our approach is to train another convolutional network that captures useful aspects of the prior keypoint distribution (not conditioned on the image). We train it to predict the position of a random subsets of keypoints, given the position of the other keypoints. More specifically, we train the convolutional network as a denoising model, similar to the denoising auto-encoder \cite{vincent2008extracting} by completely corrupting the location of a randomly chosen subset of the keypoints 
and learning to accurately predict their correct location given that of the other keypoints.
This network receives as input, not the image, but only keypoint locations represented as one-hot 2D maps (one 2D map per keypoint, with a 1 at the position of the keypoint and zeros elsewhere). It is composed of convolutional layers with large receptive fields (to get to see nearby keypoints), ReLU nonlinearities and no subsampling (see Figure \ref{fig:joint}). The network outputs probability maps for the location of all keypoints, however, its training criterion uses only prediction errors of the corrupted ones. The cost being optimized similar to Eq.(\ref{eqn:training}) but includes only the corrupted keypoints.


%
%
%
%

Once, this denoising model is trained, the output of RCN (the predicted most likely location in one-hot binary location 2D map format) is fed to the denoising model. We then simply sum the pre-softmax values of both RCN and denoising models and pass them through a softmax to generate the final output probability maps.
The joint model is depicted in Figure \ref{fig:joint}. 
The joint model combines the RCN's predicted conditional distribution for keypoint $k$ given the image $P(Y_k | X = x)$
with the denoising model's distribution of the location of that keypoint given other keypoints $P(Y_{k} | Y_{\neg{k}})$,
to yield an estimation of keypoint $k$'s location given both image and other keypoint locations  $P(Y_{k} | Y_{\neg{k}}, X = x)$.
The choice of convolutional networks
for the denoising model allows it to be easily combined with RCN in a unified deep convolutional architecture.

\section{Experimental setup and results}
\label{experiment}
We evaluate our model\footnote{Our models and code are publicly available at https://github.com/SinaHonari/RCN} on the following datasets with evaluation protocols defined by previous literature:

\textbf{AFLW and AFW datasets:}
Similar to TCDCN \cite{zhang2014facial}, we trained our models on the MTFL dataset,\footnote{MTFL consists of 10,000 training
images: 4151 images from LFW \cite{huang2007labeled} and 5849 images from the web.} which we split into
9,000 images for training and 1,000 for validation.
We evaluate our models on the same 
subsets of 
AFLW \cite{kostinger2011annotated} and AFW \cite{zhu2012}
used by \cite{zhang2014facial}, consisting of 
2995 and 377 images, respectively, each labeled with 5 facial keypoints. \todo{mention here how the datasets are quite different, making this problem especially hard! Give a pointer to the section where we tried to solve this through NADE/RNADE/augmentation/occlusion.}

\textbf{300W dataset:}
300W \cite{sagonas2013300} standardizes multiple datasets into one common dataset with 68 keypoints.
The training set is composed of 3148 images (337 AFW, 2000 Helen, and 811 LFPW). The test
set is composed of 689 images (135 IBUG, 224 LFPW, and 330 Helen). The IBUG is referred to as the
challenging subset, and the union of LFPW and Helen test sets is referred to as the common subset. We shuffle the training set and split it into
90\% train-set (2834 images) and 10\% valid-set (314 images). 

One challenging issue in these datasets is that the test set examples are significantly different and more difficult compared to the
training sets. In other words the train and test set images are not from the same distribution. In particular,
the AFLW and AFW test sets contain many samples with occlusion and more extreme rotation and expression cases than the training set.
The IBUG subset of 300W
contains more extreme pose and expressions than other subsets.

\textbf{Error Metric:}
The euclidean distance between the true and estimated landmark positions 
normalized by the distance between the eyes (interocular distance) is used:

\vspace*{-.6em}
{\footnotesize
\begin{eqnarray}
\textrm{error} = \frac{1}{K N}\sum_{n=1}^{N}\sum_{k=1}^{K}\frac{\sqrt{(w^{(n)}_{k} - {\tilde{w}}^{(n)}_{k})^{2} + (h^{(n)}_{k} -{\tilde{h}}^{(n)}_{k})^{2}}}{D^{(n)}},
\label{error_metric}
\end{eqnarray}
}
where $K$ is the number of keypoints, $N$ is the total number of images, $D^{(n)}$ is the interocular distance in image $n$.
$(w^{(n)}_{k}, h^{(n)}_{k})$ and $({\tilde{w}}^{(n)}_{k}, {\tilde{h}}^{(n)}_{k})$ represent the true
and estimated coordinates for keypoint $k$ in image $n$, respectively.

\subsection{Evaluation on SumNet and RCN}
\label{sec:evaluation}

\todo{tweak wording (perhaps together)}
  
We evaluate RCN on the 5-keypoint test sets.
To avoid overfitting and improve performance,
we applied online data augmentation to the 9,000 MTFL train set using random scale, rotation, and translation jittering\footnote{We jittered data separately in each epoch, whose parameters were uniformly sampled in the following ranges (selected based on the validation set performance): 
Translation and Scaling: [-10\%, +10\%] of face bounding box size;
Rotation: [-40, +40] degrees.}.
We preprocessed images by making them gray-scale and applying local contrast normalization \footnote{RGB images performed worse in our experiments.}.
%
%
In Figure \ref{fig:supp_activations}, we show a visualization of the contribution of each branch of the SumNet to the final predictions:
the coarsest layer provides robust but blurry keypoint locations,
while the finest layer
gives detailed face information but suffers from many false positives.
However, the sum of branches in SumNet and the finest branch in RCN make precise predictions.

\begin{figure*}[h]
\begin{center}
\begin{subfigure}{1.0\textwidth}
\centering
\includegraphics[width=1\textwidth]{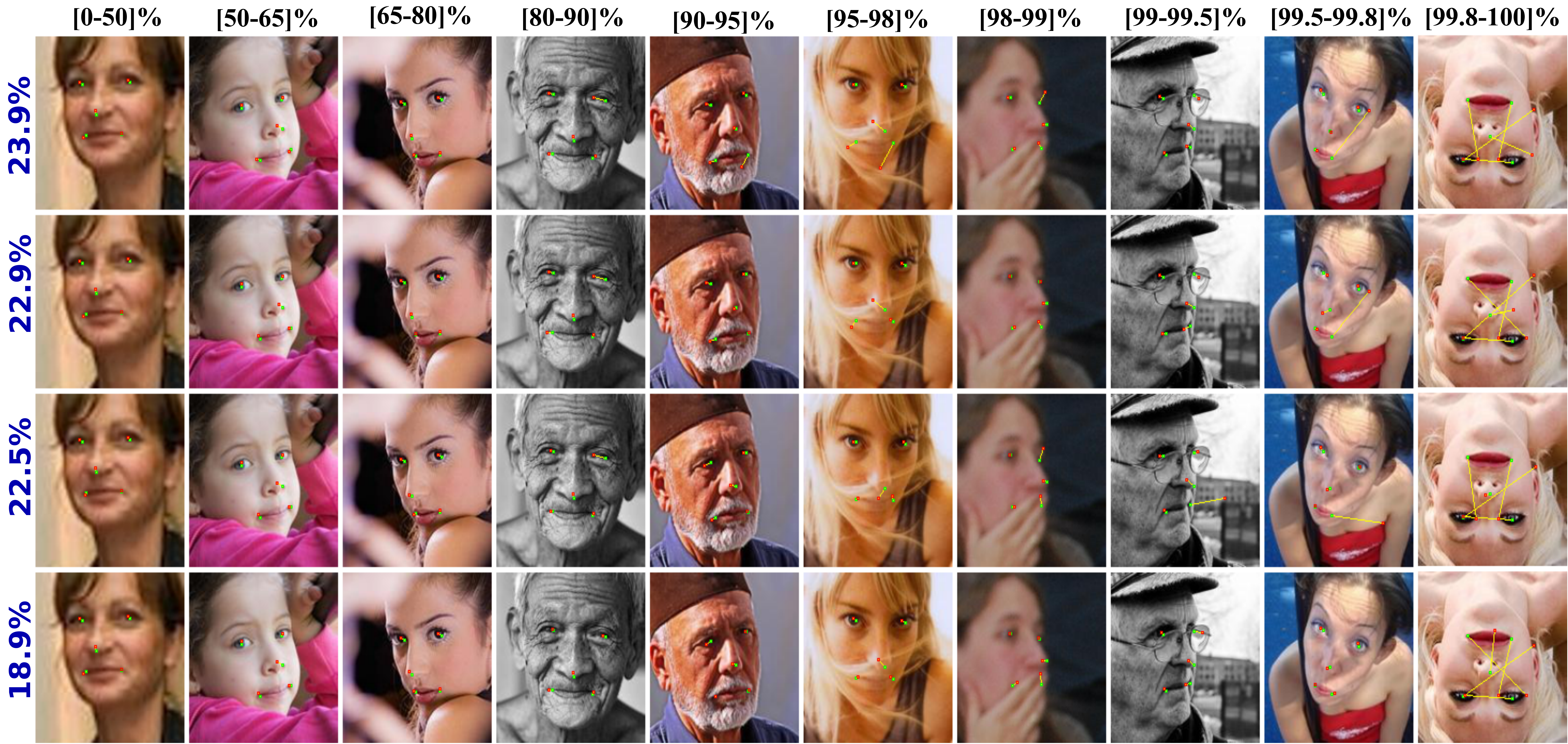}
\end{subfigure}
\end{center}
\caption{
Keypoint predictions on random test set images from easy (left) to hard (right).
Each column shows predictions of following models from top to bottom:
SumNet, SumNet with occlusion, RCN, RCN with occlusion (all models have 5 branches).
We note for each test set image (including both AFLW and AFW) the average error over the four models and use this as a notion of that image's difficulty. We then sort all images by difficulty and draw a random image from percentile bins, using the bin boundaries noted above the images.
To showcase the models' differing performance, we show only a few easier images on the left side and focus more on the hardest couple percent of images toward the right side.
The value on the left
is the average error of these samples per model (much higher than the results reported in
Table~\ref{tab:compare_MTFL}
because of the skew toward difficult images). The yellow line connects the true keypoint location (green)
to the model's prediction (red). Dots are small to avoid covering large regions of the image. Best viewed with zoom in color.
Figure~\ref{fig:supp_comp_difficulty_order} shows the performance of these four models as the difficulty of the examples increase.
}
\label{fig:cf_compare}
\end{figure*}

\begin{figure*}[h]
\begin{center}
\scalebox{.96}{
\hspace*{-.7em}
\begin{subfigure}{1.04\textwidth}
\centering
\fboxsep=0mm
\fboxrule=1pt
\fcolorbox{green}{yellow}
{\includegraphics[width=0.285\textwidth]{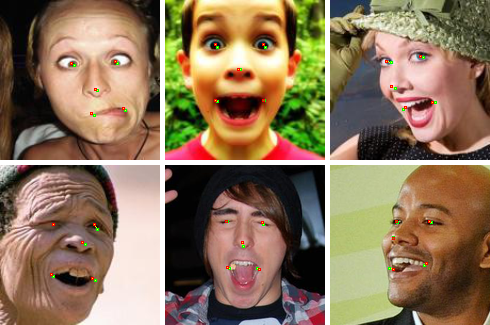}}
\hspace{.2em}
\fboxsep=0mm
\fboxrule=1pt
\fcolorbox{red}{yellow}
{\includegraphics[width=0.285\textwidth]{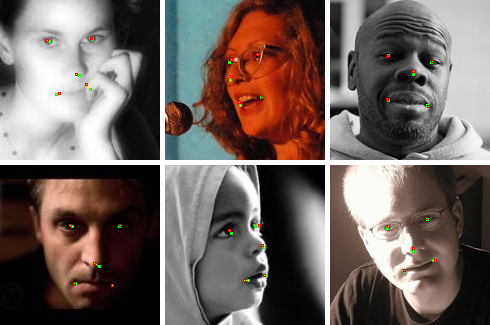}}
\hspace{.2em}
\fboxsep=0mm
\fboxrule=1pt
\fcolorbox{blue}{yellow}
{\includegraphics[width=0.38\textwidth]{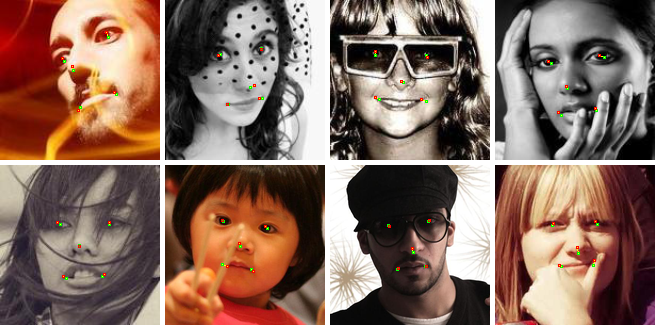}}
\end{subfigure}
}
\end{center}
\caption{Samples with different expressions (green border), contrast and illuminations (red border) and occlusions (blue border) from AFLW and AFW sets. In each box,
top row depicts samples from SumNet and bottom row shows samples from RCN, both with occlusion pre-processing. }
\label{fig:model_samples}
\end{figure*}

\begin{figure*}[htb]
\begin{center}
\begin{subfigure}{1.0\textwidth}
\centering
\includegraphics[width=0.95\textwidth]{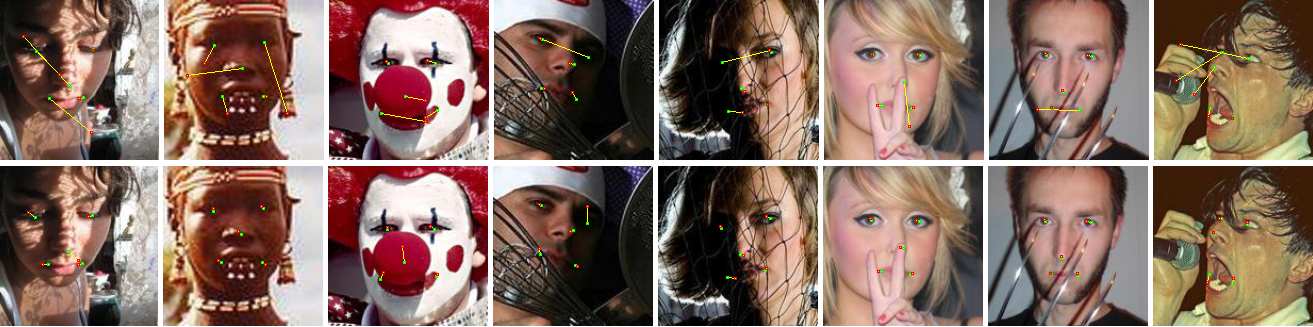}
\end{subfigure}
\end{center}
\caption{Samples from AFLW and AFW test sets showing keypoint detection accuracy without (top row) and with (bottom row) occlusion pre-processing using RCN.
}
\label{fig:occlusion}
\end{figure*}

Since the test sets contain more extreme occlusion and lighting conditions compared to the train set,
we applied a preprocessing to the train set to bring it closer to the test set distribution.
In addition to the jittering, we found it helpful to occlude images in the training set with randomly placed black rectangles\footnote{Each image was occluded with one black (zeros) rectangle, whose size was drawn uniformly in the range [20, 50] pixels.
It's location was drawn uniformly over the entire image.} at each training iteration.
This trick forced the convnet models to use more global facial components to localize the keypoints and not rely as much on the features around
the keypoints, which in turn, made it more robust against occlusion and lighting contrast in the test set.
Figure \ref{fig:cf_compare} shows the effects of this occlusion when used to train the SumNet and RCN models on randomly drawn samples.
The samples show for most of the test set examples the models do a good prediction.
Figure \ref{fig:model_samples} shows some hand-picked examples from the test sets, to show extreme expression, occlusion and contrast that
are not captured in the random samples of Figure \ref{fig:cf_compare}.
Figure \ref{fig:occlusion} similarly uses some manually selected examples to show the benefits of using occlusion.

To evaluate how much each branch contributes to the overall performance of the model, 
we trained models excluding some branches and report the results in Table \ref{tab:mask}.
The finest 
layer on its own does a poor job due to many false positives, while the coarsest layer on its own does a reasonable job, but still lacks high accuracy.
One notable result is that using only the coarsest and
finest branches together produces reasonable performance.
However, the best performance is achieved by using all branches, merging four resolutions of coarse, medium, and fine information.
\begin{table}[th]
\begin{center}
\vspace*{-.1em}
\resizebox{0.75\linewidth}{!}{
\begin{tabular}{c||c|c||c|c}
 & \multicolumn{2}{c||}{SumNet} & \multicolumn{2}{c}{RCN} \\
\hline
Mask & AFLW & AFW & AFLW & AFW \\
\hline

1, 0, 0, 0 & 10.54 & 10.63 & 10.61 & 10.89\\

0, 1, 0, 0 & 11.28 & 11.43 & 11.56 & 11.87\\

1, 1, 0, 0 & 9.47 & 9.65 & 9.31 & 9.44\\

0, 0, 1, 0 & 16.14 & 16.35 & 15.78 & 15.91\\

0, 0, 0, 1 & 45.39 & 47.97 & 46.87 & 48.61 \\

0, 0, 1, 1 & 13.90 & 14.14 & 12.67 & 13.53\\

0, 1, 1, 1 & 7.91 & 8.22 & 7.62 & 7.95\\

1, 0, 0, 1 & 6.91 & 7.51 & 6.79 & 7.27\\

\textbf{1, 1, 1, 1} & \textbf{6.44} & \textbf{6.78} & \textbf{6.37} & \textbf{6.43} 
\end{tabular}
}
\end{center}
\caption{The performance of SumNet and RCN trained
with masks applied to different branches. 
A mask value of $1$ indicates the branch is included in the model and $0$ indicates it is omitted (as a percent; lower is better).
In SumNet model mask $0$ indicates no contribution from that branch to the summation of all branches, while in RCN, if a branch is omitted, the previous coarse branch is upsampled to the following fine branch. The mask numbers are ordered from the coarsest branch to the finest branch.}
\label{tab:mask}
\end{table}
We also experimented with adding extra branches, getting to a coarser resolution of 5 $\times$ 5 in the 5 branch model, 2 $\times$ 2 in the 6 branch model 
and 1 $\times$ 1 in the 7 branch model. In each branch, the same number of convolutional layers with the same kernel size is applied,\footnote{A single exception is that when the 5 $\times$ 5 resolution map is reduced to 2 $\times$ 2,
we apply 3 $\times$ 3 pooling with stride 2 instead of the usual
2 $\times$ 2 pooling, to keep the resulting map left-right symmetric.}
and all new layers have 48 channels.
The best 
performing model, as shown in Table \ref{tab:convnets}, is RCN with 6 branches.
Comparing RCN and SumNet training, RCN converges faster. Using early stopping and without occlusion pre-processing, RCN requires on average 200 epochs to converge (about 4 hours on a NVidia Tesla K20 GPU), while SumNet needs on average more than 800 epochs (almost 14 hours).
RCN's error on both test sets drops below 7\% on average after only 15 epochs (about 20 minutes), while
SumNet needs on average 110 epochs (almost 2 hours) to get to this error.
Using occlusion preprocessing increases these times slightly but results in lower test error.
At test time, a feedforward pass on a K20 GPU takes 2.2ms for SumNet and 2.5ms for RCN per image
in Theano \cite{bastien2012theano}.
Table \ref{tab:convnets} shows occlusion pre-processing significantly helps boost the accuracy of RCN, while slightly helping SumNet.
We believe this is due to global information flow from coarser to finer branches in RCN.
\begin{table}[ht]
\begin{center}
\resizebox{0.8\linewidth}{!}{
\begin{tabular}{l||c|c}
Model & AFLW & AFW \\
\hline
SumNet (4 branch) & 6.44 & 6.78 \\
SumNet (5 branch) & 6.42 & 6.53 \\
SumNet (6 branch) & 6.34 & 6.48 \\
SumNet (5 branch - occlusion) & 6.29 & 6.34  \\
SumNet (6 branch - occlusion) & 6.27 & 6.33  \\
RCN (4 branch) & 6.37 & 6.43\\
RCN (5 branch) & 6.11 & 6.05\\
RCN (6 branch) & 6.00 & 5.98\\
RCN (7 branch) & 6.17 & 6.12\\
RCN (5 branch - occlusion) & 5.65 & 5.44 \\
RCN (6 branch - occlusion) & \textbf{5.60} & \textbf{5.36} \\
RCN (7 branch - occlusion) & 5.76 & 5.55 \\
RCN (6 branch - occlusion - skip) & 5.63 & 5.56 \\
\end{tabular}
}
\end{center}
\caption{SumNet and RCN performance with different number of branches, occlusion preprocessing and skip connections.}
\label{tab:convnets}
\end{table}
\addtocounter{footnote}{-1}
\begin{table}[t] 
\newcolumntype{d}[1]{D{.}{.}{#1}}
\centering
\resizebox{0.9\linewidth}{!}{
\begin{tabular}{c|c|c}
Model & \multicolumn{1}{c|}{AFLW} & \multicolumn{1}{c}{AFW} \\
\hline

TSPM \cite{zhu2012} & 15.9 & 14.3 \\

CDM \cite{yu2013} & 13.1 & 11.1 \\

ESR \cite{cao2014} & 12.4 & 10.4 \\

RCPR \cite{burgos2013} & 11.6 & 9.3 \\

SDM \cite{xiong2013} & 8.5 & 8.8 \\

TCDCN \cite{zhang2014facial} & 8.0 & 8.2 \\

\shortstack{TCDCN baseline (our implementation)} & 7.60 & 7.87 \\

SumNet (FCN/HC) baseline (this) & 6.27 & 6.33\\

RCN (this) & \multicolumn{1}{c|}{\textbf{5.60}} & \multicolumn{1}{c}{\textbf{5.36}} \\

\end{tabular}
}
\caption{Facial landmark mean error normalized by interocular distance on AFW and AFLW sets (as a percent; lower is better).\protect\footnotemark}
\label{tab:compare_MTFL}
\vspace*{-.8em}
\end{table}
\footnotetext{SumNet and RCN models are trained
using occlusion preprocessing.\label{note1}}
\subsection{Comparison with other models}
\label{ex:compare}

\begin{figure*}[h]
\begin{center}
\begin{subfigure}{1.0\textwidth}
\centering
\fboxsep=0mm
\fboxrule=1pt
\fcolorbox{yellow}{yellow}
{\includegraphics[width=0.10\textwidth]{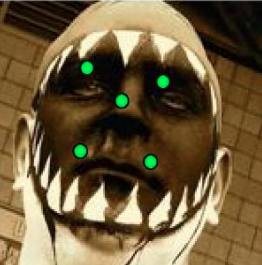}}
\fboxsep=0mm
\fboxrule=1pt
\fcolorbox{orange}{yellow}
{\includegraphics[width=0.10\textwidth]{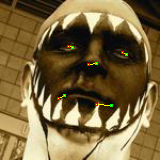}}
\fboxsep=0mm
\fboxrule=1pt
\fcolorbox{blue}{yellow}
{\includegraphics[width=0.10\textwidth]{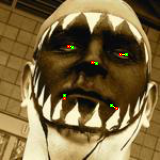}}
\hspace{.2em}
\fboxsep=0mm
\fboxrule=1pt
\fcolorbox{yellow}{yellow}
{\includegraphics[width=0.10\textwidth]{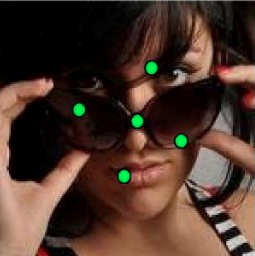}}
\fboxsep=0mm
\fboxrule=1pt
\fcolorbox{orange}{yellow}
{\includegraphics[width=0.10\textwidth]{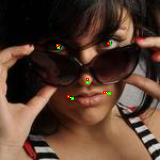}}
\fboxsep=0mm
\fboxrule=1pt
\fcolorbox{blue}{yellow}
{\includegraphics[width=0.10\textwidth]{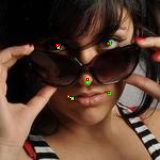}}
\hspace{.2em}
\fboxsep=0mm
\fboxrule=1pt
\fcolorbox{yellow}{yellow}
{\includegraphics[width=0.10\textwidth]{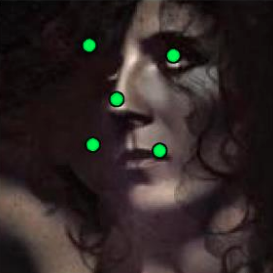}}
\fboxsep=0mm
\fboxrule=1pt
\fcolorbox{orange}{yellow}
{\includegraphics[width=0.10\textwidth]{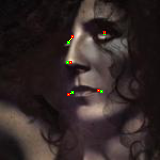}}
\fboxsep=0mm
\fboxrule=1pt
\fcolorbox{blue}{yellow}
{\includegraphics[width=0.10\textwidth]{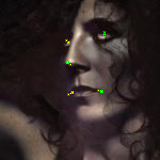}}
\end{subfigure}
\end{center}
    \caption{Samples from TCDCN \cite{zhang2014facial} (yellow border with green predicted points) versus SumNet (orange border) and RCN (blue border). In the latter two models, red and green dots show predicted and true keypoints. TCDCN samples are taken directly from \cite{zhang2014facial}.
}
\label{fig:TCDCN_compare}
\end{figure*}

\textbf{AFLW and AFW datasets:}
We first re-implemented the TCDCN model \cite{zhang2014facial},
which is the current state of the art model on 5 keypoint AFLW \cite{kostinger2011annotated} and AFW \cite{zhu2012} sets,
and applied the same pre-processing as our other experiments. Through hyper-parameter search, we
even improved upon the AFLW and AFW results reported in \cite{zhang2014facial}.
Table \ref{tab:compare_MTFL} compares RCN with other models. Especially, it improves the SumNet baseline,
which is equivalent to FCN and Hypercolumn models, and it also converges faster.
The SumNet baseline is also provided by this paper and to the best of our knowledge this
is the first application of any such coarse-to-fine convolutional architecture to the facial keypoint problem.
Figure \ref{fig:TCDCN_compare} compares TCDCN with SumNet and RCN models, on some difficult samples reported in \cite{zhang2014facial}.

\begin{figure*}[h]
\begin{center}
\begin{subfigure}{1.0\textwidth}
\centering
\includegraphics[width=1.0\textwidth]{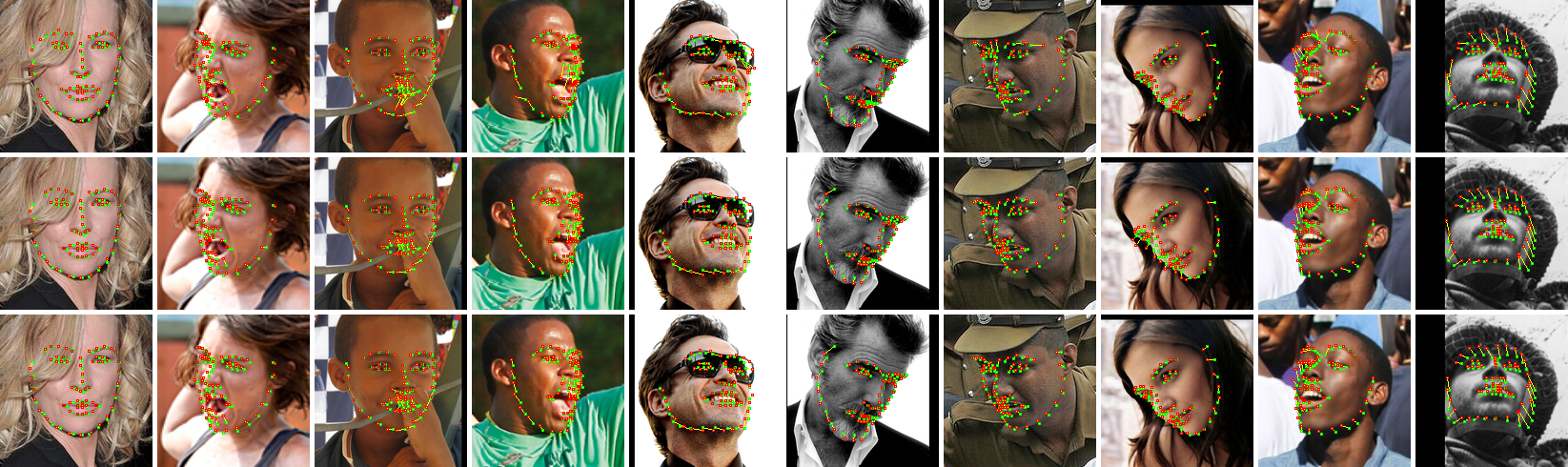}
\end{subfigure}
\end{center}
\caption{Samples from 300W test sets. Each column shows samples in this order (top to bottom): RCN, keypoint denoising model and the joint model.
The first two columns show extreme expression and occlusion samples where RCN's prediction is highly accurate. The next 5 columns
show samples where the denoising model improves the RCN's predictions. In the 8th column the structured model find a
reasonable keypoint distribution but deteriorates the RCN's predictions.
Finally, the last two columns show cases where the denoising
model generates plausible keypoint distributions but far from the true keypoints.}
\label{fig:compare_300W}
\end{figure*}

\textbf{300W dataset \cite{sagonas2013300}:}
The RCN model that achieved the best result on the validation set,
contains 5 branches with 64 channels for all layers (higher capacity is needed to extract features for more keypoints) and 2 extra
convolutional layers with $1\times1$ kernel size in the finest branch right before applying the softmax.
Table \ref{tab:300W_compare}
compares different models on all keypoints (68) and a subset of keypoints (49) reported in \cite{tzimiropoulos2015project}.
The denoising model is trained by
randomly choosing 35 keypoints
in each image and jittering them
(changing their location uniformly to any place in the 2D map).
It improves the RCN's prediction by considering
how locations of different keypoints are inter-dependent. Figure \ref{fig:compare_300W}
compares the output of RCN, the denoising model and the joint model, showing how the keypoint distribution
modeling can reduce the error.
We only trained RCN on the
2834 images in the train-set. No extra data is taken to pre-train or
fine-tune the model \footnote{We only jittered the train-set images by random scaling, translation and rotation similar to the 5 keypoint dataset. \\
${\quad}^{\dagger}$ TCDCN \cite{zhang2015learning} uses 20,000 extra dataset for pre-training.}.
The current state-of-the-art model without any extra data${}^{\dagger}$ is CFSS\cite{zhu2015face}. We reduce the error by 15\% on the IBUG subset compared to CFSS.

\begin{table}[t]
\newcolumntype{d}[1]{D{.}{.}{#1}}
\centering
\resizebox{1.\linewidth}{!}{
\begin{tabular}{c|c|c|c|c}
Model & \multicolumn{1}{c|}{\#keypoints} & \multicolumn{1}{c|}{Common} & \multicolumn{1}{c|}{IBUG} & \multicolumn{1}{c}{Fullset} \\
\hline
PO-CR \cite{tzimiropoulos2015project} &  & 4.00 & 6.82 & 4.56 \\

RCN (this) & 49 & 2.64 & 5.10 & 3.88 \\

\shortstack{RCN + denoising \\keypoint model (this)} & & \textbf{2.59} & \textbf{4.81} & \textbf{3.76} \\

\hline
CDM \cite{yu2013} & & 10.10 & 19.54 & 11.94 \\

DRMF \cite{asthana2013robust} &  & 6.65 & 19.79 & 9.22 \\

RCPR \cite{burgos2013} &  & 6.18 & 17.26 & 8.35 \\

GN-DPM \cite{tzimiropoulos2014gauss} & & 5.78 & - & -\\

CFAN \cite{zhang2014coarse} & & 5.50 & 16.78 & 7.69 \\

ESR \cite{cao2014} &  & 5.28 & 17.00 & 7.58 \\

SDM \cite{xiong2013} & 68 & 5.57 & 15.40 & 7.50 \\

ERT \cite{cao2014face} & & - & - & 6.40 \\

LBF \cite{ren2014face} & & 4.95 & 11.98 & 6.32 \\

CFSS\cite{zhu2015face} & & 4.73 & 9.98 & 5.76 \\

$\mathrm{TCDCN\,}^{\dagger}$ \cite{zhang2015learning} &  & 4.80 & 8.60 & 5.54 \\

RCN (this) & & 4.70 & 9.00 & 5.54 \\

\shortstack{RCN + denoising \\keypoint model (this)}  & & \textbf{4.67} & \textbf{8.44} & \textbf{5.41} \\

\end{tabular}
}
\caption{Facial landmark mean error normalized by interocular distance on 300W test sets (as a percent; lower is better). \textsuperscript{\ref{note1}}}
\label{tab:300W_compare}
\vspace*{-1em}
\end{table}
\section{Conclusion}
In this paper we have introduced the Recombinator Networks architecture for combining coarse maps of pooled features with fine non-pooled features in convolutional neural networks. The model improves upon previous summation-based approaches by feeding coarser branches into finer branches, allowing the finer resolutions to learn upon the features extracted by coarser branches. We find that this new architecture leads to both reduced training time and increased facial keypoint prediction accuracy.
We have also proposed a denoising model for keypoints which involves explicit modeling of valid spatial configurations of keypoints. This allows our complete approach to deal with more complex cases such as those with  occlusions.
%
%
\subsubsection*{Acknowledgments}
We would like to thank the Theano developers, particularly F. Bastien and P. Lamblin, 
for their help throughout this project. 
We appreciate Y. Bengio and H. Larochelle feedbacks and also L. Yao, F. Ahmed and M. Pezeshki's helps in this project.
We also thank
Compute Canada, and Calcul Quebec for providing computational resources. 
Finally, we would like to thank
Fonds de Recherche du Qu\'ebec -- Nature et Technologies
(FRQNT) for a doctoral
research scholarship (B2) grant during 2014 and 2015 (SH) and the NASA Space Technology Research Fellowship (JY).
\definecolor{goodclr}{rgb}{0,.8,0}
\begin{table*}[h]
\centering
\begin{tabular}{c||c|c|c|c|c}

\backslashbox{Features}{Models} & \shortstack{Efficient \\Localization \cite{tompson2014efficient}} & \shortstack{Deep \\ Cascade \cite{sun2013deep}} & \shortstack{Hyper-\\columns \cite{hariharan2014hypercolumns}} & \shortstack{FCN \\ \cite{long2014fully}} & \shortstack{RCN \\ (this)} \\
\hline

\shortstack{Coarse features: hard crop or soft combination?} & {\color{red}Hard} & {\color{red}Hard} & {\color{goodclr}Soft} & {\color{goodclr}Soft} & {\color{goodclr}Soft} \\
\hline

\shortstack{Learned coarse features fed into finer branches?} & \shortstack{\color{red}No} &  \shortstack{\color{red}No} &
\shortstack{\color{red}No} & \shortstack{\color{red}No} & \shortstack{\color{goodclr}Yes} \\

\end{tabular}
\caption{Comparison of multi-resolution architectures. The Efficient Localization and Deep Cascade models use coarse features
to crop images (or fine layer features),
which are then fed into fine models. This process saves computation when dealing with high-resolution
images but at the expense of making a greedy decision halfway through the model. Soft models merge local and global features of the entire image
and do not require a greedy decision.
The Hypercolumn and FCN models propagate all coarse information to the final layer but merge information via addition instead of conditioning fine features on coarse features. The Recombinator Networks (RCN), in contrast, injects coarse features directly into finer branches, allowing the fine computation to be tuned by (conditioned on) the coarse information. The model is trained end-to-end and results in \emph{learned} coarse features which are tuned directly to support the eventual fine predictions.
  \todo{Explain ``greedy early crop'' vs. ``soft final decision''. Also mention how a greedy crop can be good if you want to save computation.}
  \maybe{What should people conclude from this? Tell them what to think.}
  \maybe{However, when dealing with high resolution images, the approaches in \cite{sun2013deep} and \cite{tompson2014efficient} reduce computational cost, since the fine convnets at the end of the pipeline are forced to work on a patch.}
}
\label{tab:models}
\end{table*}

%
%
%
%

\clearpage 
{\small
\bibliographystyle{ieee}
\bibliography{references}
}

\clearpage 

\setcounter{equation}{0}
\setcounter{figure}{0}
\setcounter{table}{0}
\setcounter{page}{1}
\makeatletter
\renewcommand{\theequation}{S\arabic{equation}}
\renewcommand{\thefigure}{S\arabic{figure}}
%
%
\maketitle
%
%
\begin{figure*}[!htb]
\centering
\textbf{\Large Supplementary Information for \\Recombinator Networks: Learning Coarse-to-Fine Feature Aggregation\\}
\vspace{2.0em}
\begin{subfigure}{0.02\textwidth}
\caption{}
\label{fig:c1}
\end{subfigure}
\scalebox{0.99}{
\begin{minipage}[b]{0.88\textwidth}
\includegraphics[width=0.54\textwidth]{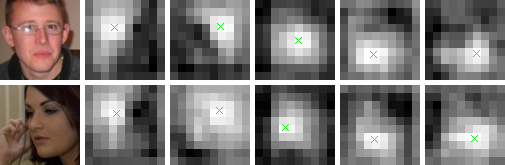}
\includegraphics[width=0.45\textwidth]{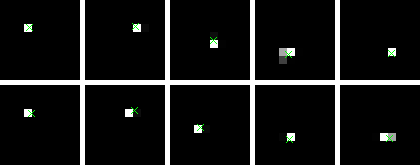}
\end{minipage}
}
\vspace{-0.3em}
\\
\begin{subfigure}{0.02\textwidth}
\caption{}
\label{fig:c2}
\end{subfigure}
\scalebox{0.99}{
\begin{minipage}[b]{0.88\textwidth}
\includegraphics[width=0.54\textwidth]{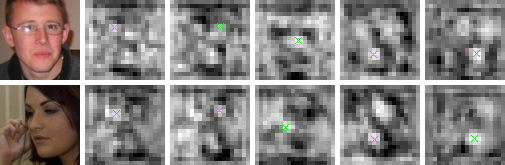}
\includegraphics[width=0.45\textwidth]{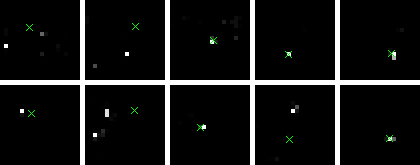}
\end{minipage}
}
\vspace{-0.3em}
\\
\begin{subfigure}{0.02\textwidth}
\caption{}
\label{fig:c3}
\end{subfigure}
\scalebox{0.99}{
\begin{minipage}[b]{0.88\textwidth}
\includegraphics[width=0.54\textwidth]{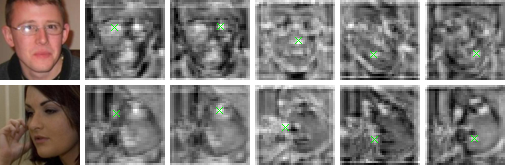}
\includegraphics[width=0.45\textwidth]{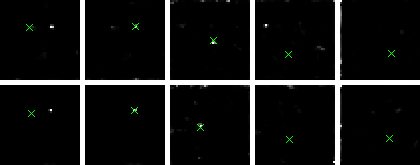}
\end{minipage}
}
\vspace{-0.3em}
\\
\begin{subfigure}{0.02\textwidth}
\caption{}
\label{fig:c4}
\end{subfigure}
\scalebox{0.99}{
\begin{minipage}[b]{0.88\textwidth}
\includegraphics[width=0.54\textwidth]{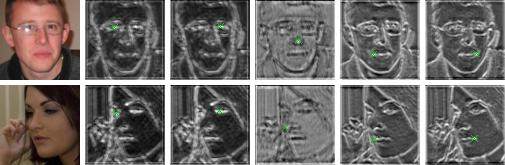}
\includegraphics[width=0.45\textwidth]{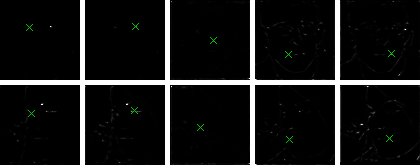}
\end{minipage}
}
\vspace{-0.3em}
\\
\begin{subfigure}{0.02\textwidth}
\caption{}
\label{fig:sum}
\end{subfigure}
\scalebox{0.99}{
\begin{minipage}[b]{0.88\textwidth}
\includegraphics[width=0.54\textwidth]{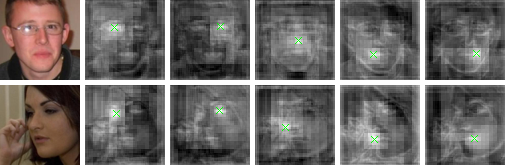}
\includegraphics[width=0.45\textwidth]{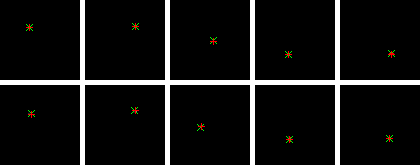}
\end{minipage}
}
\vspace{-0.3em}
\\
\begin{subfigure}{0.015\textwidth}
\caption{}
\label{fig:sum_chain}
\end{subfigure}
\scalebox{0.99}{
\begin{minipage}[b]{0.88\textwidth}
\includegraphics[width=0.54\textwidth]{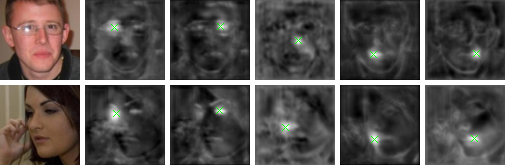}
\includegraphics[width=0.45\textwidth]{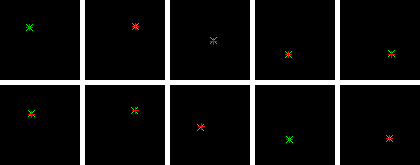}
\end{minipage}
}
\caption{
    Sub-figures \subref{fig:c1}, \subref{fig:c2}, \subref{fig:c3},  \subref{fig:c4} show pre-sum (left) and softmax (right) of the
    coarsest to finest branches in a 4-branch SumNet model. The softmax used in these branches are only for illustration purposes
    and is not part of the trained model.
    Sub-figure \subref{fig:sum} (left) shows the sum of branches in the SumNet model and Sub-figure \subref{fig:sum_chain} (left) depicts the pre-softmax values in RCN.
    The true keypoint locations are shown by green cross in all figures to show their relative correspondence with the branch activations.
    SumNet and RCN's predictions are shown by red plus on the post-softmax maps in Sub-figures \subref{fig:sum} (right) and \subref{fig:sum_chain} (right), respectively.
    In each row the images correspond to the keypoints in this order from left to right: left-eye, right-eye, nose, left-mouth, right-mouth. Best viewed electronically with zoom.
    }
\label{fig:supp_activations}
\end{figure*}
\clearpage

\begin{minipage}{\textwidth}
\centering
\includegraphics[width=.6\textwidth]{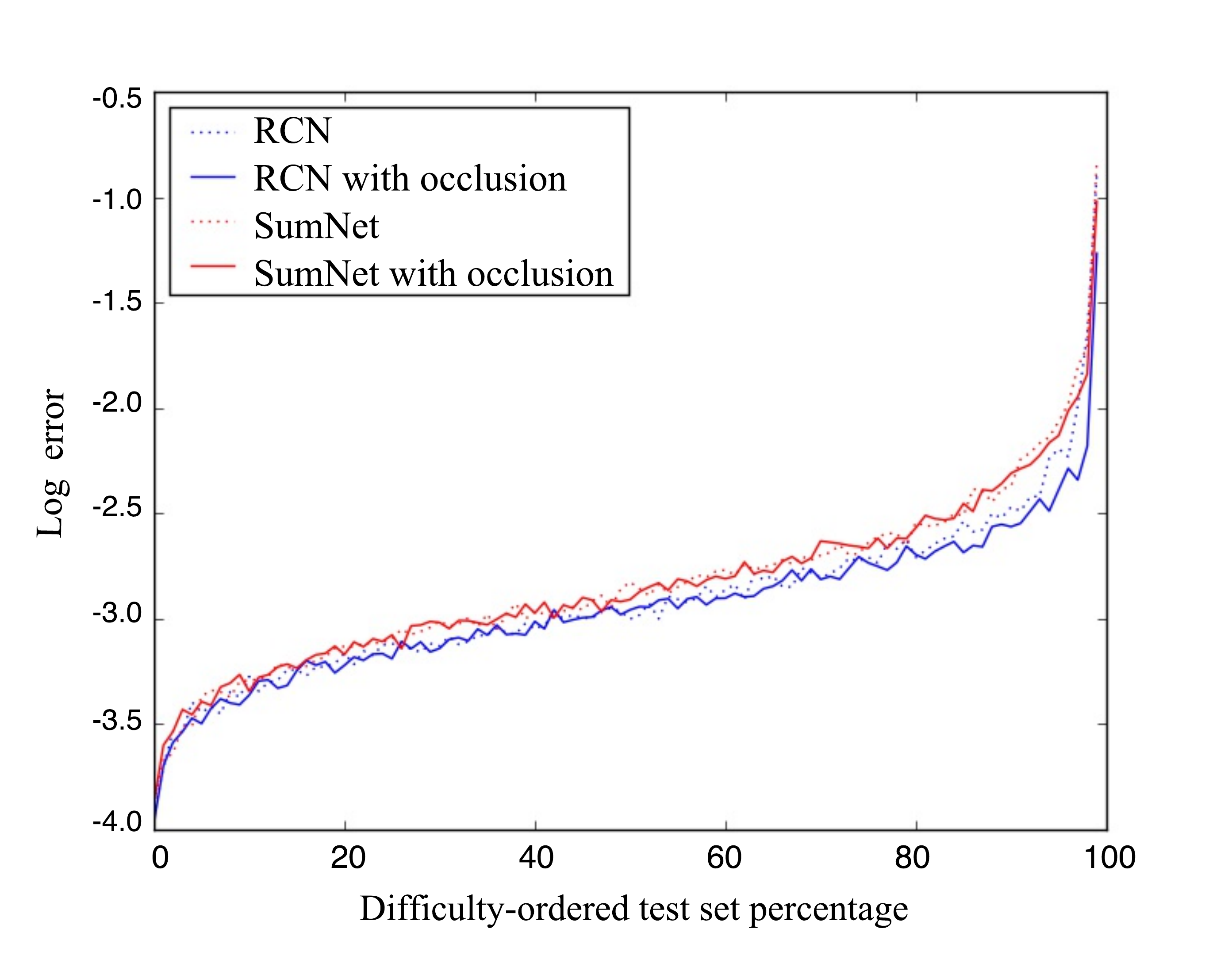}
\captionof{figure}{The performance of SumNet and RCN models with and without occlusion pre-processing on the merged AFW and AFLW test sets as the difficulty of the examples increase (lower is better).
To get this plot, we note for each test set image (including both AFLW and AFW) the average error over the four models and use this as a notion of that image's difficulty. We then sort all images by difficulty and get each model's log error (using Eq.~\ref{error_metric}) on each test example.
Finally, we plot each model's performance on the sorted test set examples from the easiest (0\% difficulty) to the most difficult (100\% difficulty) percentage of the test set examples.
The plot shows RCN performs better than SumNet, especially on the harder examples. The occlusion pre-processing helps RCN on most difficult examples (difficulty $>$ 65\%), while it slightly helps SumNet.}
\label{fig:supp_comp_difficulty_order}
\end{minipage}

\end{document}